\documentclass{article}

\usepackage[final,nonatbib]{neurips_2020}

\usepackage[utf8]{inputenc} 
\usepackage[T1]{fontenc}    
\usepackage{hyperref}       
\usepackage{url}            
\usepackage{booktabs}       
\usepackage{amsfonts}       
\usepackage{nicefrac}       
\usepackage{microtype}      

\usepackage{amsthm}
\theoremstyle{plain}
\makeatletter
\newtheorem*{rep@theorem}{\rep@title}
\newcommand{\newreptheorem}[2]{%
\newenvironment{rep#1}[1]{%
 \def\rep@title{#2 \ref{##1}}%
 \begin{rep@theorem}}%
 {\end{rep@theorem}}}
\makeatother
\newtheorem{thm}{\protect\theoremname}
\providecommand{\theoremname}{Theorem}
\newreptheorem{thm}{Theorem}
\newtheorem{cor}{\protect\corollaryname}
\providecommand{\corollaryname}{Corollary}
\newreptheorem{cor}{Corollary}
\newtheorem{lem}{\protect\lemmaname}
\providecommand{\lemmaname}{Lemma}
\newreptheorem{lem}{Lemma}
\usepackage{cancel}

\usepackage{microtype}
\usepackage{graphicx}
\usepackage{booktabs} 

\usepackage{times}
\usepackage{graphicx} 
\usepackage{algorithm}
\usepackage{algorithmic}
\usepackage{amsmath}
\usepackage[normalem]{ulem}
\usepackage{caption}
\usepackage{subcaption}
\usepackage{multirow}
\usepackage{tabularx}
\usepackage{enumitem}
\usepackage{verbatim}

\usepackage{hyperref}

\title{Learning Individual Interactions from Population Dynamics with Discrete-Event Simulation Model}

\author{%
  Yan Shen\\
  Department of Computer Science and Engineering\\
  University at Buffalo, NY.\\
  \texttt{yshen22@buffalo.edu} \\
   \And
   Fan Yang \\
   Department of Computer Science and Engineering\\
    University at Buffalo, NY.\\
  \texttt{fyang24@buffalo.edu} \\
   \texttt{email} \\
   \AND
   Mingchen Gao \\
   Department of Computer Science and Engineering\\
    University at Buffalo, NY.\\
  \texttt{mgao8@buffalo.edu} \\
   \texttt{email} \\\
   \And
   Wen Dong \\
   Department of Computer Science and Engineering\\
    University at Buffalo, NY.\\
  \texttt{wendong@buffalo.edu} \\
   \texttt{email} \\
}

\begin{document}

\maketitle

\begin{abstract}

The abundance of data affords researchers to pursue more powerful computational tools to learn the dynamics of complex system, such as neural networks, engineered systems and social networks. Traditional machine learning approaches capture complex system dynamics either with dynamic Bayesian networks and state space models, which is hard to scale because it is non-trivial to prescribe the dynamics with a sparse graph or a system of differential equations; or a deep neural networks, where the distributed representation of the learned dynamics is hard to interpret. In this paper, we will explore the possibility of learning a discrete-event simulation representation of complex system dynamics assuming multivariate normal distribution of the state variables, based on the observation that many complex system dynamics can be decomposed into a sequence of local interactions, which individually change the system state only minimally but in sequence generate complex and diverse dynamics. Our results show that the algorithm can data-efficiently capture complex network dynamics in several fields with meaningful events.

\end{abstract}

\section{Introduction}

We live in a complex world, where the complex interactions of a large number of elementary units at the microscopic level give rise of the complex system behavior at the macroscopic level. Examples of such complex systems include neural activities in our brain, gene regularization and signal transduction networks in systems biology, the movement of people in an urban system, and epidemic and opinion dynamics in social networks. In learning and making inferences about the dynamics of such systems, we can gain valuable insights in optimizing their operations, for example about functional areas of the brain and relevant diagnoses, about traffic congestion and more efficient use of roads, and about how people are infected in an epidemic crisis. The plummeting cost and increasing availability of computational resources have enabled researchers to identify explainable models of complex system behavior through data-driven discovery methods, for example in the brain activities measured with functional magnetic resonance imaging (fMRI) and electroencephalogram (EEG), people's movement and interactions tracked with ubiquitous computing technologies, and the diffusion and mutation of opinions in social media.

In this paper, we will derive an algorithm that learns a discrete-event simulation (DES) representation of complex system dynamics from complete and noisy observations, assuming multivariate normal distribution of the state variables. The hypothesis is that many complex system dynamics can be decomposed into a sequence of local interactions, which individually change the system state only minimally but in sequence generate complex and diverse dynamics. Discrete-event simulation is a versatile framework with diverse applications in capturing the dynamics of gene regularization network using stochastic kinetic model, engineered systems using Petri net model, social systems using stock and flow chart. To learn the discrete-event simulation representation, we will use a generalized expectation maximization algorithm to jointly learn a collection of local events whose rates (counts per unit time) are sparsely dependent the state variables and a stoichiometric matrix that mixes the event counts into the state variable changes. We expect the learned discrete-event simulator to enjoy a high level of machine learning performance in predicting unstudied dynamics from training data, and at the same time provide useful insights on why the trained model behaves in a specific way and an intuitive interface to correct any errors produced by the model.

One approach to represent a multivariate time-series is through state-space models, where the time-derivative of state variables is identified as the sparse interactions among them. Along this line of research, sparse regression methods were developed to discover a system of partial differential equations from spatio-temporal measurements of the system \cite{brunton2016discovering,rudy2017data}. Symbolic regression and evolutionary methods were used to synthesize and select the state-space representations of data that balance model accuracy and complexity \cite{bongard2007automated,schmidt2009distilling}. The plummeting cost of sensors and rising availability of computational resources has enabled many deep learning approaches in data-driven discovery methods including \cite{pascanu2013construct,chang2016compositional,raissi2018deep,wang2019towards,ma2022fewshot}. With a graph neural network, a state-space model is represented as a directed graph, where nodes represent state variables and edges represent their relations. The state evolution in a graph neural network is simulated with messages passing from nodes to edges and messages from edges to nodes, both emulated using deep neural networks \cite{battaglia2016interaction,battaglia2018relational}. In comparison with these parameterized approaches to model the deterministic state evolution, we use a different inductive basis through assuming that the state evolution is driven by a collection of stochastic events each involving the interactions of a few system components.

Another approach to model the structure of a multivariate time-series is through Gaussian processes. In linear coregionalization analysis in geostatistics \cite{alvarez2012kernels}, and its machine learning counterparts such as semiparametric latent factor model \cite{teh2005semiparametric}, multi-task Gaussian process \cite{bonilla2008multi}, and Gaussian process emulator \cite{o2006bayesian}, a vector-valued random function is expressed as a linear combination of independent scalar-valued Gaussian processes of predictor variables. These vector-valued random functions of predictor variables can be stacked into an expressive deep Gaussian process state-transition kernel of a autoregressive process and learned through variational inference \cite{mattos2015recurrent}. Other Gaussian process approaches include identifying a low-dimensional embedding of a high-dimensional state-space model where the predictor-dependent loading matrix is expressed as a sparse combination of Gaussian process dictionary functions \cite{fox2015bayesian}, and identifying independent clusters in a high-dimensional state-space model \cite{ren2017clustering} with a Dirichlet process mixture model. In comparison with the non-parametric and semi-parametric Gaussian process approaches, we learn a parameterized discrete-event representation of a state-space model where the state change is induced by a collection of independent events each involving the interactions of a few state variables.

A discrete event model captures the macroscopic dynamics of a complex system through a sequence of stochastic events which each involve the interactions of a few individuals in the system microscopically and in aggregate induce complex dynamics. The sparse interactions thus introduces an inductive prior in the representation of complex system dynamics. The discrete event model is used to specify the dynamics of engineered systems (where it is known as a stochastic Petri-net \cite{marsan1994modelling}), biochemical networks (where it is known as a stochastic kinetic model \cite{wilkinson2011stochastic}), and social networks (where it is known as discrete event simulation \cite{borshchev2013big}), and random variables (where it is know as a continuous time Bayesian network \cite{nodelman2002continuous}). Existing works to learn and make inferences about discrete-event dynamics generally assume that the interactions are known, and proceed to learn the probabilities for these interactions to happen. These works involve making approximate inferences with Markov chain Monte Carlo \cite{golightly2011bayesian,rao2013fast}, structured mean field \cite{opper2008variational,zhang2017collapsed}, expectation propagation \cite{xu2016using} and deep neural networks \cite{tran2017deep}. In comparison, we aim to learn a collection of mutually independent local interactions to capture complex system dynamics as a distributed but explainable representation, where each interaction happens with a probability that depends on and changes the states of a few system variables.


\section{Discrete Event Representation of Dynamical Systems}

A variety of models to specify complex system dynamics are based on the premise that the complex dynamics can be factored into a sequence of sparse interactions of the system's elementary individuals. Such systems include the stochastic kinetic model, the stochastic Petri-net, the discrete-event simulator, and the continuous-time Bayesian network. Hence, an algorithm to learn complex system dynamics in terms of mutually independent local interactions assumes a reasonable prior on the dynamics and is applicable to diverse complex systems.

A \textbf{stochastic kinetic model} is a biochemist's way of describing the temporal evolution of a biochemical network with $M$ species driven by $V$ mutually independent events, where the stochastic effects are particularly prevalent (e.g., a transcription network or signal transduction network). Let $\mathbf X=(\mathbf X^{(1)},\cdots,\mathbf X^{(M)})$ denote individuals belonging to the $M$ species in the network. An event (chemical reaction) $v$ is specified by a production
\begin{align} 
 & { \alpha_{v}^{(1)}\mathbf X^{(1)}+\cdots+\alpha_{v}^{(M)}\mathbf X^{(M)}\overset{c_v}\to\beta_{v}^{(1)}\mathbf X^{(1)}+\cdots+\beta_{v}^{(M)}\mathbf X^{(M)}}.\label{eq:production}
\end{align}
The production is interpreted as having \emph{rate constant} $c_{v}$ (probability per unit time, as time goes to 0), $\alpha_{v}^{(1)}$ individuals of species $1$ ... interact according to event $v$, resulting in their removal from the system; and $\beta_{v}^{(1)}$ individuals of species $1$ ... are introduced into the system. Hence event $v$ changes the populations by $\Delta_{v}=(\beta_{v}^{(1)}- \alpha_{v}^{(1)},\cdots,\beta_{v}^{(M)}-\alpha_{v}^{(M)})^\top$, and $S=(\Delta_1,\cdots,\Delta_V)$ is the \emph{stoichiometric matrix}. A stochastic kinetic process can be simulated through the Gillespie algorithm \cite{gillespie2007stochastic}, which defines a compound Poisson process measure to a sample path induced by a sequence of events.

A \textbf{Petri net} (or place/transition net) is a model of a discrete-event dynamic system in the area of systems-theory and automatic-control, with applications in modeling and optimizing manufacturing logistic, computer systems and engineered systems \cite{marsan1994modelling}. A Petri net is defined as a 4-tuple $(S, T, W M_0)$, where $S$ is a finite set of \emph{places} (equivalent to species in stochastic kinetic model), $T$ is a set of \emph{transitions} (equivalent to events), $W:(S\times T)\cup(T\times S)$ is a \emph{multi-set of arcs} (equivalent to the stoichiometricity of reactants $(\alpha_{i,j})$ and products $(\beta_{i,j})$), $M$ is a \emph{marking} that assigns a number of \emph{tokens} (equivalent to individuals) to each place and $M_0$ is the \emph{initial marking}. In a Petri net, a transition $t\in T$ may \emph{fire} in a marking $M$ if there are enough tokens in its input places ($\forall s: M(s)\ge W(s,t)$), and firing a transition in $M$ consumes $W(s,t)$ tokens from input places $s$ and produces $W(t,s)$ tokens in output places $s$. 

A \textbf{system dynamics model} is a tool to understanding and controlling the nonlinear behaviour of complex systems using stocks, flows and internal feedback loops, with applications in policy research in the areas of corporate management, urban dynamics and economical systems. The basis of this approach is that the behavior of a complex system is often determined by the interaction structure of the system's components. In a system dynamics model, stocks are analogous to species, flows are analogous to events/transitions, the positive and negative feedback loops implement the control of a system, and the dynamics is expressed as difference/differential equations and simulated with a computer. 

\begin{figure}

\begin{subfigure}[b]{0.15\textwidth}\includegraphics[width=1\columnwidth]{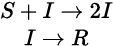}
\caption{\label{fig:ProductionRule}Production rule system}
\end{subfigure}\hfill
\begin{subfigure}[b]{0.20\textwidth}\includegraphics[width=1\columnwidth]{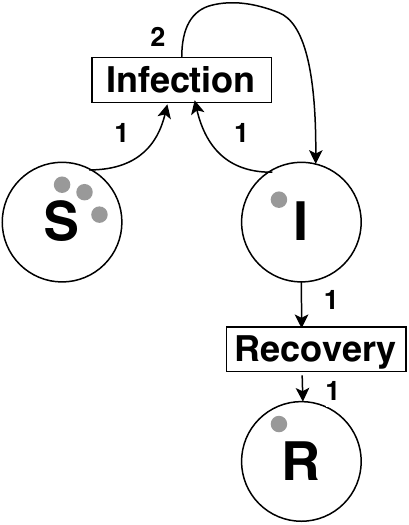}
\caption{\label{fig:PetNet}Petri-net}
\end{subfigure}\hfill
\begin{subfigure}[b]{0.30\textwidth}\includegraphics[width=1\columnwidth]{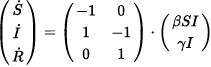}
\caption{\label{fig:PDE}PDE system}
\end{subfigure}\hfill
\begin{subfigure}[b]{0.25\textwidth}\includegraphics[width=1\columnwidth]{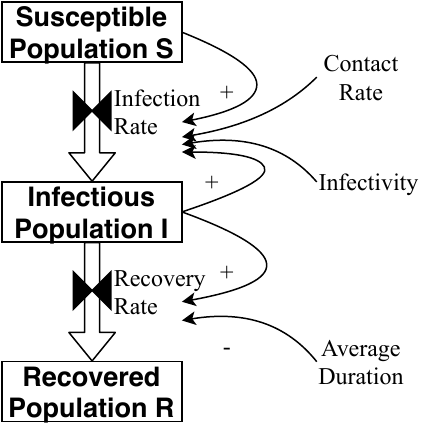}
\caption{\label{fig:StockFlow}Stock and flow diagram}
\end{subfigure}\hfill
\caption{\label{fig:Models} Several representations of a discrete event process. (a) Production rule system, (b) Petri net. (c) System of partial differential equations. (d) Stock and flow diagram.}
\end{figure} 

Fig. \ref{fig:Models} shows several representations of the susceptible-infectious-recovered model, where the production rule system and the Petri-net capture individual level dynamics, and the system of partial differential equations and the stock and flow diagram capture population level dynamics. While the representations come from different disciplines, they are all parameterized by the stoichiometric matrix and the event rate constants, suggesting their significance when we learn the dynamics of many complex interaction systems from data. 

There are many other models to capture complex dynamics through a collection of local interactions. For example, a compound Poisson process specifies the dynamics through a sequence of event times and a discrete-time Markov process conditioned on the event times. A continuous time Bayesian network captures the dyadic interactions among the nodes through events happening on edges. A multiagent simulator specifies the dynamics through a set of events --- infection and recovery in an epidemic simulator such as EpiSims, entering and exiting a road link in a road transportation simulator such as MATSim, and user-defined events in a general purpose simulator such as AnyLogic and NetLogo. In the following, we will formulate the event-learning problem in a Gaussian dynamical system and derive an algorithm to solve this problem.

\section{Methodology}

In this section, we formulate the problem of learning the interactions of the elementary units in a discrete event process at the microscopic level to explain the discrete-time observations of population dynamics at the macroscopic level. With this formulation, we construct computational graphs to learn the discrete event representation through establishing the equivalence relationship between gradient ascent and generalized expectation maximization.

\subsection{Problem Definition}

The learning problem is to identify a minimum set of production rules (Eq. \ref{eq:production}) with event rate constants $\mathbf c=(c_1,...,c_V)\in[0,\infty)^V$ and sparse multiplicity factors $\alpha_v=(\alpha_v^{(1)},...,\alpha_v^{(M)})\in[0,\infty)^M$ and $\beta_v=(\beta_v^{(1)},...,\beta_v^{(M)})\in[0,\infty)^M$ to fit the time-discretized Langevin dynamics to the population dynamics of the system observed at discrete times. Specifically, we are solving the parameter estimation problem in Eq. \ref{eq:problem} , where random vector $\mathbf x_{t}=(x_{t}^{(1)},\dots,x_{t}^{(M)})\in[0,\infty)^M$ represents the populations of the \textit{M} species in the system at time $t$, $\mathbf y_t$ represents observation about $\mathbf x_t$ at time $t$, vector-valued function $\mathbf h(\mathbf x_t)=\left(h_1(\mathbf x_t),\dots,h_V(\mathbf x_t)\right)\in[0,\infty)^V$ represents the rates of \textit{V} independent interactions among the system's elementary units, real-valued matrix $S$ is the stoichiometric matrix that ``mixes'' the numbers of interaction events into state changes, and $\Delta t$ is an infinitesimal time step to linearize the (nonlinear) Langevin dynamics.

\begin{align}
 & \operatornamewithlimits{arg min}_{\mathbf c,\alpha,\beta} -\log p(\mathbf y_{t_1},...,\mathbf y_{t_N})\!+\!\!\!\sum_v\! \lambda_{\alpha,1} \|\alpha_v\|_1\!+\!\!\lambda_{\alpha,2}\|\alpha_v\|_2^2 \!+\!\!\lambda_{\beta,1} \|\beta_v\|_1\!+\!\!\lambda_{\beta,2}\|\beta_v\|_2^2\!+\!\!\lambda_{\mathbf c}\|\mathbf  c\|_1,\label{eq:problem}\\
 & \mathbf x_{t_n} = \mathbf x_{t_{n-1}} + S\mathbf h(\mathbf x_{t_{n-1}})\Delta t + S\operatorname{diag}\sqrt{\mathbf h(\mathbf x_{t_{n-1}})\Delta t}\cdot\omega_{t_n},\label{eq:DSKM-state-model}\\
& \mathbf y_{t_n}=\mathbf y(\mathbf x_{t_n}, \nu_{t_n}),\label{eq:DSKM-observation-model}\\
& \mathbf h(\mathbf x_{t_n}) = \exp\left(\log (\mathbf c) +  \left(\alpha_1,\dots,\alpha_V\right)^\top\cdot\log\mathbf x_{t_n}\right),\label{eq:h}\\
& S = \left(\beta_1-\alpha_1,...,\beta_V-\alpha_V\right)^\top\label{eq:S},\\
& \omega_{t_n}\stackrel{i.i.d.}{\sim}\mathcal N(0,I), \nu_{t_n}\stackrel{i.i.d.}{\sim}\mathcal N(0,I), t_n=n\Delta t,\mathbf c\in[0,\infty)^V,\alpha_v,\beta_v\in[0,\infty)^M.\nonumber
\end{align}

Generally the observations are made at a coarser temporal resolution than the necessary resolution to realistically simulate the Langevin dynamics. Without loss of generality, we assume that the observations are made at the fine time resolution $t_{1}=\Delta t,t_{2}=2\cdot\Delta t,\dots$, and set  $p(\mathbf y_{t_{n}}|\mathbf x_{t_{n}})=1$ when no actual observation is received. In the following, we use $\mathbf x_1,\mathbf x_2,\dots,$ and $\mathbf y_1,\mathbf y_2,\dots,$ interchangeably with $\mathbf x_{t_1},\mathbf x_{t_2},\dots,$ and $\mathbf y_{t_1},\mathbf y_{t_2}$, and so on.

The Langevin dynamics \cite{Golightly2007Diffusion} is a Gaussian approximation of the Gillespie algorithm \cite{gillespie2007stochastic}, which simulates the discrete-event dynamics starting from $\mathbf x_0$ at time $t=0$ by iteratively (1) sampling the time $\tau$ to the next event according to exponential distribution $\tau\sim\operatorname{Exp}(h_0(\mathbf x_t;\mathbf c))$, where $h_0(\mathbf x_t;\mathbf c)=\sum_{v=1}^{V}h_{v}(\mathbf x_t,c_{v})$ (2) sampling the event $v_{t+\tau}$ according to categorical distribution $v_{t+\tau}\sim\operatorname{Cat}\left(\frac{h_{1}}{h_{0}},\dots,\frac{h_{V}}{h_{0}}\right)$, and (3) updating the populations $x_{t+\tau}\leftarrow x_t+\beta_v-\alpha_v$ and system time $t\leftarrow t+\tau$. In the Gaussian approximation, the event counts in the interval $[t,t+\Delta t)$ are approximately Poisson and mutually independent, and can be further approximated with a multivariate normal random variable $\mathbf r_{t+\Delta t}\sim\operatorname{Poi}(\mathbf h(\mathbf x_t)\Delta t)\approx\mathcal N(\mathbf h(\mathbf x_t)\Delta t,\operatorname{diag}\sqrt{\mathbf h(\mathbf x_t)\Delta t})$. The formulation of event rates (Eq. \ref{eq:h}) has an interpretation in the mass-action stochastic kinetics literature, where $c_{v}$ is the probability per unit time for $\alpha^{(m)}_v$ selected individuals from species $m=1,\cdots,M$ to interact according to event $v$, and $(x^{(m)}_t)^{\alpha^{(m)}_v}$ is the total number of ways to select $\alpha^{(m)}_v$ molecules out of $x^{(m)}_t$ molecules belong to species $m$ assuming homogeneous populations throughout the configuration space or in an infinitesimal volume. The total event rate regardless of which molecules we select is thus $c_v\prod_m (x^{(m)}_t)^{\alpha^{(m)}_v}$. The Langevin dynamics is a reflected stochastic process  \cite{veestraeten2004conditional} where the populations are required to be non-negative. Here we use data away from the boundary to make the learning algorithm and its analysis conceptually simple.

Comparing with graph neural network \cite{battaglia2016interaction}, Langevin dynamics (Eq. \ref{eq:DSKM-state-model}) similarly quantify the interactions among system variables on the edges connecting the corresponding system variable nodes, and update the system variables from aggregating the quantities on the incident edges of the nodes. Different from graph neural network, Langevin dynamics provide a way to learn hyper-edges (in contrast with dyads in GNN) through learning $\alpha_v$s and $\beta_v$s in a Bayesian framework, and the system variables are specially interpreted as populations.

The following theorem says that we can simulate any diffusion process from the Langevian dynamics defined in Eq. \ref{eq:DSKM-state-model} with a proper specification of populations under mild conditions. Thus despite the seeming restrictions on non-negative populations and event driving dynamics, the Langevian dynamics is quite expressive.

\begin{thm}\label{thm:existence}
Let $d \mathbf x_{t}=\mathbf \mu(\mathbf x_t) dt + \sqrt{\Sigma(\mathbf x_t)} d \mathbf B_t$, where $B_t$ is a Brownian motion process, $\mu(\mathbf x_t)$ is a drift term and $\Sigma(\mathbf x_t)$ is a covariance matrix. Let $\max_{\mathbf x}\|\operatorname{diag}(\sqrt{|\mu(\mathbf x)|})\|_2\le C_1$ and $\min_{\mathbf x}\|\sqrt{\Sigma(\mathbf x)}\|_2\ge C_2$. Then $\mathbf x_t$ can be written in the form of Langevian dynamics Eq. \ref{eq:DSKM-state-model}.
\end{thm}

\begin{proof}[Proof sketch]
We write $\mathbf x_t = \mathbf x_t^+ - \mathbf x_t^- $, where $\mathbf x_t^+=\max(\mathbf x_t, 0)$ and $\mathbf x_t^-=-\min(\mathbf x_t, 0)$. Next we define populations as $(\mathbf z_t^+, \mathbf z_t^-)$, where $\mathbf z_t^+ = C \mathbf x_t^+$, $\mathbf z_t^- = C \mathbf x_t^-$, and $C$ is a constant satisfying $C\ge C_1/C_2$. The populations thus induce a new stochastic process $\mathbf z_t = \mathbf z_t^+ - \mathbf z_t^- = C \mathbf x_t$ with dynamics $d \mathbf z_{t}=C  \mu(\frac{1}{C}(\mathbf z_t^+-\mathbf z_t^-)) dt + \sqrt{C^2\Sigma(\frac{1}{C}(\mathbf z_t^+-\mathbf z_t^-))} d \mathbf B_t$, satisfying $C^2\Sigma-\operatorname{diag}(C|\mu|)$ is positive definite (because for all $\mathbf w\ne 0$, $\mathbf w^\top (C^2\Sigma) \mathbf w - \mathbf w^\top (C\operatorname{diag}|\mu|) \mathbf w \ge (C_2 C^2-C_1  C)\|\mathbf w\|_2^2>0$). Then we define events each involving one population to account for the drift term $\mu(\mathbf x_t)$. Each such event increases or decreases the corresponding population by 1 according to $\operatorname{sign}(C \mu(\frac{1}{C}(\mathbf z_t^+-\mathbf z_t^-)))$ with rate $C \left|\mu(\frac{1}{C}(\mathbf z_t^+-\mathbf z_t^-))\right|$. Then we define event pairs to account for the remaining diffusion $C^2\Sigma-\operatorname{diag}(C|\mu|)$, with each pair involving two populations. The pair of events change the two populations $i$ and $j$ with rate $C|\Sigma_{i,j}(\frac{1}{C}(\mathbf z_t^+-\mathbf z_t^-))|$, by $(1,1)$ and $(-1,-1)$ if $\operatorname{sign}(C\Sigma_{i,j}))=1$ and by $(1,-1)$ and $(-1,1)$ if $\operatorname{sign}(C\Sigma_{i,j}))=-1$. Detailed proof is in the supplementary material.
\end{proof}

\subsection{Learning Discrete-Event Simulation Dynamics from Discrete-Time Observations}

In this section, we construct a gradient descent algorithm to learn the discrete-event dynamics through minimizing the negative log-likelihood of a sequence of discrete-time observations about the latent state. 

One potential issue with gradient-based optimization is the vanishing/exploding gradient problem when negative log likelihood is constructed iteratively over time involving a deep stack of function compositions. The following theorem shows that a gradient descent algorithm to minimize the negative log likelihood of observations is a generalized expectation maximization algorithm to maximize the expected log-likelihood. The reason is that the derivative of the observation log-likelihood over model parameters equals the expected value of the derivative of the complete data (observations + latent state) log-likelihood over the posterior distribution of the latent state conditioned on observations. As a result, we can construct a computational graph to estimate the negative log likelihood of observations, and expect the back propagation algorithm to minimize negative log likelihood to have the same behavior as a generalized expectation maximization algorithm.

\begin{thm}
\label{thm:gradient}Let $p(x,y;\theta)$ be the joint probability
distribution of latent variable $x$ and observable variable $y$ parameterized
by $\theta$. Then $\nabla_{\theta}\log p(y;\theta)=\mathbf{E}_{p(x|y;\theta)}\nabla_{\theta}\log p(x,y;\theta)$.
\end{thm}
\begin{proof}
\begin{align*}
 & \nabla_{\theta}\log p(y;\theta)=\nabla_{\theta}\log\int dxp(x,y;\theta) = \nabla_{\theta}\int dxp(x,y;\theta)/\int dxp(x,y;\theta)\\
 =& \int dx\nabla_{\theta}p(x,y;\theta)/\int dxp(x,y;\theta) = \int dxp(x,y;\theta)\nabla_{\theta}\log p(x,y;\theta)/\int dxp(x,y;\theta)\\
=& \int dxp(x|y;\theta)\nabla_{\theta}\log p(x,y;\theta)=\mathbf{E}_{p(x|y;\theta)}\nabla_{\theta}p(x,y;\theta).
\end{align*}
\end{proof}

A corollary of Theorem \ref{thm:gradient} is that we can construct the terms in $\nabla_\theta\log p(\mathbf y_1,...,\mathbf y_N)=\sum_{n}\nabla_\theta\log p(\mathbf y_{n}|\mathbf y_1,...,\mathbf y_{n-1})$ from the posterior latent state distributions $p(\mathbf x_{n-1},\mathbf x_{n}|\mathbf y_{1},...,\mathbf y_{N})$.

\begin{cor}
\label{cor:gradient}Let $p(\mathbf x_{1},\dots,\mathbf x_{N},\mathbf y_{1},\dots,\mathbf y_{N};\theta)=\prod_{n}p(\mathbf x_{n}|\mathbf x_{n-1};\theta)p(\mathbf y_{n}|\mathbf x_{n};\theta)$ be the joint probability distribution of a latent process $\mathbf x_{1},...,\mathbf x_{N}$ and its observations $\mathbf y_{1},...,\mathbf y_{N}$ parameterized by $\theta$.
Then $\nabla_\theta\log p(\mathbf y_1,...,\mathbf y_N) = \sum_{n=1}^N\mathbf E_{p(\mathbf x_{n-1},\mathbf x_{n}|\mathbf y_1,...,\mathbf y_N)} \nabla_\theta\log p(\mathbf x_n,\mathbf y_n|\mathbf x_{n-1})$.
\end{cor}

The following technical lemma makes explicit the relationship between  $\mathbf{E}_{p(\mathbf x_{n-1},\mathbf x_t|\mathbf y_1,..,\mathbf y_N)}\nabla_{\theta}\log p(\mathbf x_{n},\mathbf y_{n}|\mathbf x_{n-1};\theta)$ and $\nabla_{\theta}\log p(\mathbf y_{n}|\mathbf y_1,...,\mathbf y_{n-1};\theta)$, and leads to a direct proof of Corollary \ref{cor:gradient} with the fact that the telescopic sum $\sum_n \mathbf E_{p(\mathbf x_n|\mathbf y_1,...,\mathbf y_N)}\nabla_\theta\log p(\mathbf x_n|\mathbf y_1,...,\mathbf y_n) - \mathbf E_{p(\mathbf x_{n-1}|\mathbf y_1,...,\mathbf y_N)}\nabla_\theta\log p(\mathbf x_{n-1}|\mathbf y_1,...,\mathbf y_{n-1}) = 0$. The direct proof of Corollary \ref{cor:gradient} through Lemma \ref{cor:gradient} is in supplementary material. 

\begin{lem} \label{thm:gradient-recursion}For $n\le m$,
\begin{align*}
& \mathbf E_{p(\mathbf x_n|\mathbf y_{1},...,\mathbf y_m)}\nabla_\theta\log p(\mathbf x_n|\mathbf y_{1},...,\mathbf y_n)\ =\  \mathbf E_{p(\mathbf x_{n-1}|\mathbf y_{1},...,\mathbf y_m)}\nabla_\theta\log p(\mathbf x_{n-1}|\mathbf y_{1},...,\mathbf y_{n-1})\\
& \hspace{.4in}+  \mathbf E_{p(\mathbf x_{n-1},\mathbf x_n|\mathbf y_{1},...,\mathbf y_m)}\nabla_\theta\log p(\mathbf x_n, \mathbf y_n|\mathbf x_{n-1};\theta) -\nabla_\theta \log p(\mathbf y_n|\mathbf y_{1},...,\mathbf y_{n-1})
\end{align*}
\end{lem}

We use the extended Kalman filter to learn the discrete-event dynamics through maximizing the log likelihood (Eqs. \ref{eq:problem} and \ref{eq:EKF-likelihood}) and estimate the latent states from noisy observations. Specifically we iteratively predict latent state distributions $p(\mathbf x_{n}|\mathbf y_{1},...,\mathbf y_{n-1})\sim\mathcal{N}( \hat{\mathbf x}_{n|n-1},P_{n|n-1})$ from $p(\mathbf x_{n-1}|\mathbf y_{1},...,\mathbf y_{n-1})$ (Eqs. \ref{eq:EKF-predicted-mean} and \ref{eq:EKF-predicted-var}), and update the predicted state distribution according to the new observations $p(\mathbf x_{n}|\mathbf y_{1},...,\mathbf y_{n})\sim\mathcal{N}(\hat{\mathbf x}_{n|n},P_{n|n})$ (Eq. \ref{eq:EKF-updated-mean-var}) in filtering pass, assuming initial state distribution $\mathbf x_{0}\sim\mathcal{N}(\hat{\mathbf x}_{0},P_{0})$. Then we iteratively predict $p(\mathbf x_{n}|\mathbf y_{1},...,\mathbf y_{N})\sim\mathcal{N}(\hat{\mathbf x}_{n|N},P_{n|N})$ from $p(\mathbf x_{n+1}|\mathbf y_{1},...,\mathbf y_{N})$ (Eq. \ref{eq:EKS-mean-var}), starting from $p(\mathbf x_{N}|\mathbf y_{1},...,\mathbf y_{N})\sim\mathcal{N}(\hat{\mathbf x}_{N|N},P_{N|N})$ in the smoothing pass. Here we use $\bullet_{n|m}$to represent the statistic about $\mathbf x_{n}$ estimated from observations $\mathbf y_{1},...,\mathbf y_{m}$, use $\hat{\mathbf x}_{n|m}$ and $P_{n|m}$ to represent the mean and variance of $\mathbf x_{n}$ conditioned on $\mathbf y_{1},...,\mathbf y_{m}$, and use notation $\partial_{\mathbf x_{n-1}}\mathbf h=\frac{\partial \mathbf h(\mathbf x_{n-1})}{\partial \mathbf x_{n-1}}|_{\mathbf x_{n-1}=\hat{\mathbf x}_{n-1|n-1}}$, $\partial_{\mathbf x_{n}}\mathbf y=\frac{\partial \mathbf y(\mathbf x_{n},\nu_{n})}{\partial \mathbf x_{n}}|_{\mathbf x_{n}=\hat{\mathbf x}_{n|n-1},\nu_{n}=0}$ and $\partial_{\nu_{n}}\mathbf y=\frac{\partial \mathbf y(x_{n},\nu_{n})}{\partial \nu_{n}}|_{\mathbf x_{n}=\hat{\mathbf x}_{n|n-1},\nu_{n}=0}$.

\begin{align}
 & \hat{\mathbf x}_{n|n-1}=\hat{\mathbf x}_{n|n-1}+S\cdot \mathbf h(\hat{\mathbf x}_{n-1|n-1},t_{n-1}),\label{eq:EKF-predicted-mean}\\
 & P_{n|n-1}=S\operatorname{diag}\left(\mathbf h(\mathbf x_{n-1},t_{n-1})\Delta t\right)S^{\top}+\left(I+S\partial_{\mathbf x_{n-1}}\mathbf h\right)P_{n-1|n-1}\left(I+S\partial_{\mathbf x_{n-1}}\mathbf h\right)^{\top},\label{eq:EKF-predicted-var}\\
  & \hat{\mathbf x}_{n|n}=\hat{\mathbf x}_{n|n-1}+K_{n}\epsilon_{n},\ \   P_{n|n}=(I-K_{n}\partial_{\mathbf x_{n}}\mathbf y)P_{n|n-1},\label{eq:EKF-updated-mean-var}\\
 & p(\mathbf y_{1},\dots,\mathbf y_{N})=\prod_{n=1}^{N}\mathcal{N}(\epsilon_{n};0,\Sigma_{\epsilon_{n}}),\label{eq:EKF-likelihood}\\
& \text{where }\epsilon_{n}=\mathbf y_{n}-\mathbf y(\hat{\mathbf x}_{n|n-1},0),\ \  \Sigma_{\epsilon_{n}}\!\!=\!\partial_{\mathbf x_{n}}\mathbf yP_{n|n-1}\partial_{\mathbf x_{n}}^{\top}\mathbf y\!+\!\partial_{\nu_{n}}\mathbf y\partial_{\nu_{n}}^{\top}\mathbf y,\ \  K_{n}=P_{n|n-1}\partial_{\mathbf x_{n}}\mathbf y\Sigma_{\epsilon_{n}}^{-1}.\nonumber \\
 & \hat{\mathbf x}_{n|N}=\hat{\mathbf x}_{n|n}+G_{n}\delta_{n},\ \   P_{n|N}=P_{n|n}+G_{n}(P_{n+1|N}-P_{n+1|n})G_{n}^{\top},\label{eq:EKS-mean-var}\\
 & \text{where }\delta_{n}=(\hat{\mathbf x}_{n+1|N}-\hat{\mathbf x}_{n+1|n}),\ \  \Sigma_{\delta_{n}}=P_{n+1|n},\ \  G_{n}\!=\!P_{n|n}\left(I+S\partial_{\mathbf x_{n}}\mathbf h\right)P_{n+1|n}^{-1}.\nonumber
\end{align}

\section{Experiments}


In this section, we compare the proposed algorithm with several state-of-the-art parametric and non-parametric algorithms in capturing complex system dynamics. The evaluation is conducted in three scenarios: predator-prey, prokaryotic auto-regulation, and SynthTown transportation network. We focus on two points. First, with less parameters, the discrete-event formulation enables us to learn dynamics faster and more data-efficiently. Second, the learned sparse interactions are more interpretable.

\subsection{Configuration for performance evaluation}

We evaluate the algorithms to learn discrete-event dynamics in three
scenarios: predator-prey, prokaryotic auto-regulation, and SynthTown. The predator-prey dynamics involve two species --- predator and prey,
and three events --- $\text{prey}\to2\text{prey}$, $\text{predator}+\text{prey}\to2\text{predator}$
and $\text{predator}\to\emptyset$. The prokaryotic auto-regulation dynamics \cite{golightly2011bayesian} involve 5 species --- DNA, P, $\text{P}_{2}$, $\text{DNA}\cdot\text{P}_{2}$, and RNA and 8 events --- $\text{DNA}+\text{P}_{2}\leftrightarrow\text{DNA}\cdot\text{P}_{2}$, $\text{DNA}\to\text{DNA}+\text{RNA}$, $\text{RNA}\to\text{RNA}+\text{P}$, $2\text{P}\leftrightarrow\text{P}_{2}$, $\text{RNA}\to\emptyset$, and $\text{P}\to\emptyset$. From
the discrete-time observations constructed with the Gillespie algorithm, we expect to recover the events. The SynthTown dynamics involve 2000 synthesized inhabitants going to work in the morning and returning home in the afternoon in a synthesized network of one home location, one work location and 23 single-direction road links specified within MATSim (a state-of-the-art transportation simulator) \cite{matsim}. From the discrete-time observations, we expect to learn events in the form $l_{j}\to l_{k}$, a vehicle moving from location $j$ to location
$k$. 

We compare the proposed algorithm with several other algorithms:
a extended Kalman filter with a general state transition dynamics,
a recurrent neural network model with LSTM cells, and a linear model
of coregionalization.

We compare the following statistics about the algorithms. First, we
compare the Kullback--Leibler divergence between the learned and
the to-be-learned dynamics with different sample sizes. To this end,
we sample trajectories $x_{1}^{(l)},\dots,x_{N}^{(l)}$ for $l=1,\dots,L$
from the learned dynamics and estimate the average log-likelihood
ratio per time step $\frac{1}{L\cdot N}\sum\limits _{l=1}^{L}\sum\limits _{n=1}^{N}\log\frac{q(x_{n}^{(l)}|x_{n-1}^{(l)})}{p(x_{n}^{(l)}|x_{n-1}^{(l)})}$.
We expect a model-based learning algorithm (such as our algorithm
to learn a discrete event model) to be more data-efficient and less
likely to experience overfit than model-free algorithms. Second, we
compare the learning speed of different algorithms. We expect Gibbs
sampling to take more time both because Gibbs sampling has a generally
slow mixing rate, and because sampling from a Gaussian process in
every Gibbs iteration involves a matrix inversion which has a cubic
time complexity in data size. All algorithms were implemented with
Keras and Tensorflow probability in R and ran on in a Linux box with
two Intel Xeon Gold 6130 CPUs and NVidia Tesla V100 GPUs each. 

We visualize the learned discrete event dynamics through comparing
how the learned dynamics and the ground truth dynamics translate event
counts into state changes, and how the two dynamics estimate the number
of events between two time steps. Since the dynamics are identical
under the permutation of events (which correspond to columns of S),
we make the comparison straightforward by setting the event rates
as $h_{v}(x)=c_{v}g_{v}(x)$, and learning event rate constants $c_{v}$
and the stoichiometric matrix $S$, thus destroying permutation invariant.

\subsection{Evaluating discrete-event structural learning}

\begin{table}
\hfill{}\begin{tabular}{|c|c|c|c|c|c|c|c|c|}
\hline 
\multirow{2}{*}{} & \multicolumn{2}{c|}{{EKF}} & \multicolumn{2}{c|}{{RNN}} & \multicolumn{2}{c|}{{GP}} & \multicolumn{2}{c|}{{DES}}\tabularnewline
\cline{2-9} \cline{3-9} \cline{4-9} \cline{5-9} \cline{6-9} \cline{7-9} \cline{8-9} \cline{9-9} 
 & {KL} & {time} & {KL} & {time} & {KL} & {time} & {KL} & {time}\tabularnewline
\hline 
\hline 
{p-p/10k} & {.2} & {2min} & {.1} & {3min} & {.04} & {10min} & {.01} & {2min}\tabularnewline
\hline 
{p-p/1m} & {.05} & {10min} & {.04} & {15min} & {---} & {---} & {.001} & {10min}\tabularnewline
\hline 
{PAG/10k} & {10} & {30min} & {7} & {50min} & {1} & {1hour} & {.1} & {20min}\tabularnewline
\hline 
{PAG (1m)} & {3} & {4hour} & {2} & {5hour} & {---} & {---} & {.01} & {2hour}\tabularnewline
\hline 
{ST/10k} & {12} & {40min} & {8} & {1hour} & {2} & {4hour} & {.2} & {30min}\tabularnewline
\hline 
{ST/1m} & {5} & {6hour} & {3} & {8hour} & {---} & {---} & {.05} & {4hour}\tabularnewline
\hline 
\end{tabular}\hfill{}

\caption{\label{tab:performance}Comparing K-L divergence and training time
of algorithms in three scenarios and with different training data
sizes. }

\end{table}

Table \ref{tab:performance} compares the achievable K-L divergence
and the corresponding training times of extended Kalman filter, recurrent
neural network, Gaussian process latent factor model and discrete-event
structural learning in learning predator-prey, prokaryotic auto-regulation,
and road traffic dynamics with training data of 10 thousand time steps
and 1 million time steps. The K-L divergence between the learned dynamics
and the ground truth dynamics is evaluated with a separate testing
trajectory. In general, learned dynamics with K-L divergence greater
than 10 are not useful, and those with K-L divergence less than 0.01
are excellent. We can easily learn useful dynamics through training
a discrete event model with a stochastic gradient algorithm and a
training data set of trajectories with 10 thousand discrete time steps
in total, while we will need 100 times more data through training
other models. Among the model-free algorithms, Gaussian process latent
factor model can achieve comparable performance as the discrete event
model, yet it is not scalable and takes longer in the training. Modeling high-dimensional dynamics with a dense neural network generally result in overfitting and doubtful learned dynamics.

\begin{figure}
\hfill{}\includegraphics[width=0.24\columnwidth]{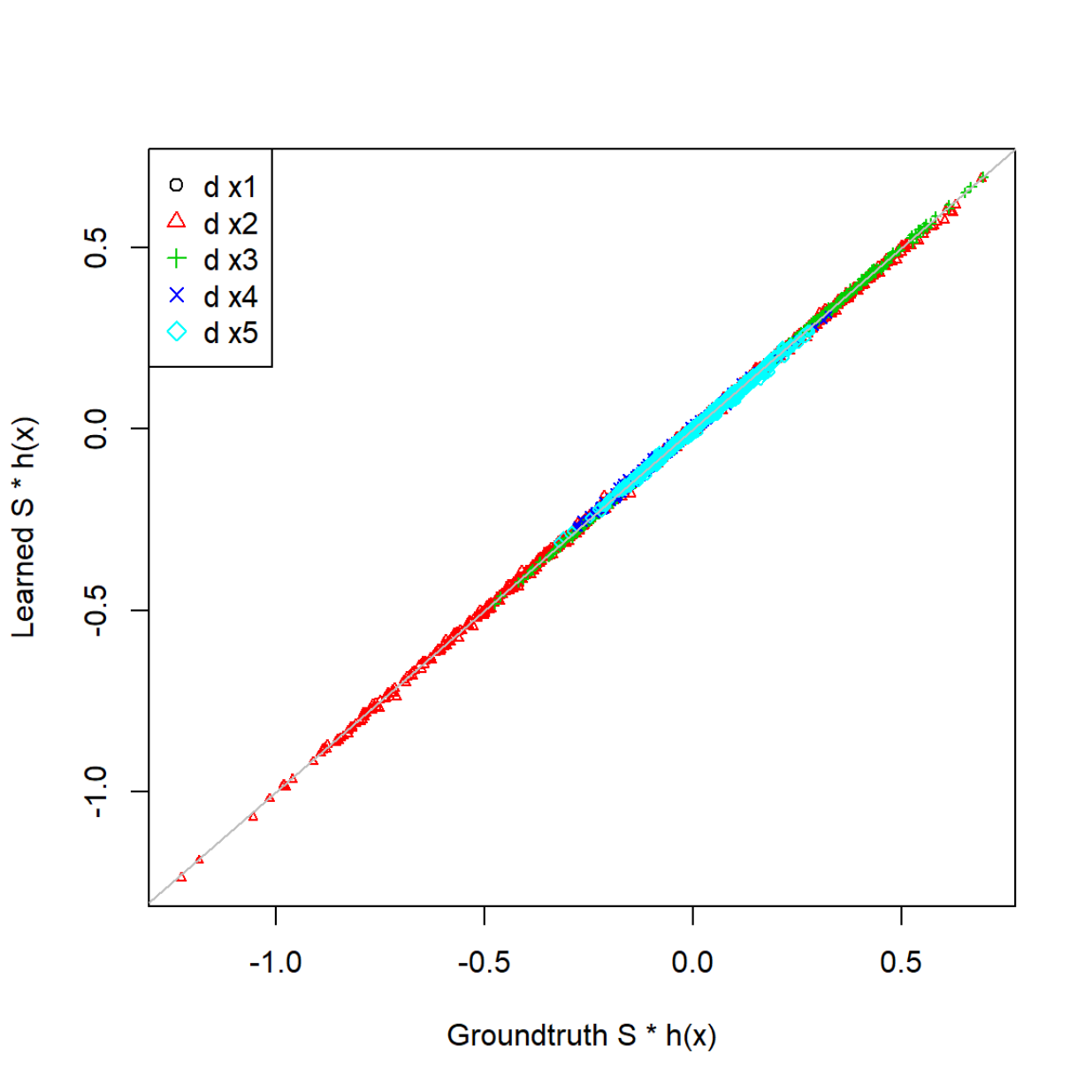}\hfill{}\includegraphics[width=0.24\columnwidth]{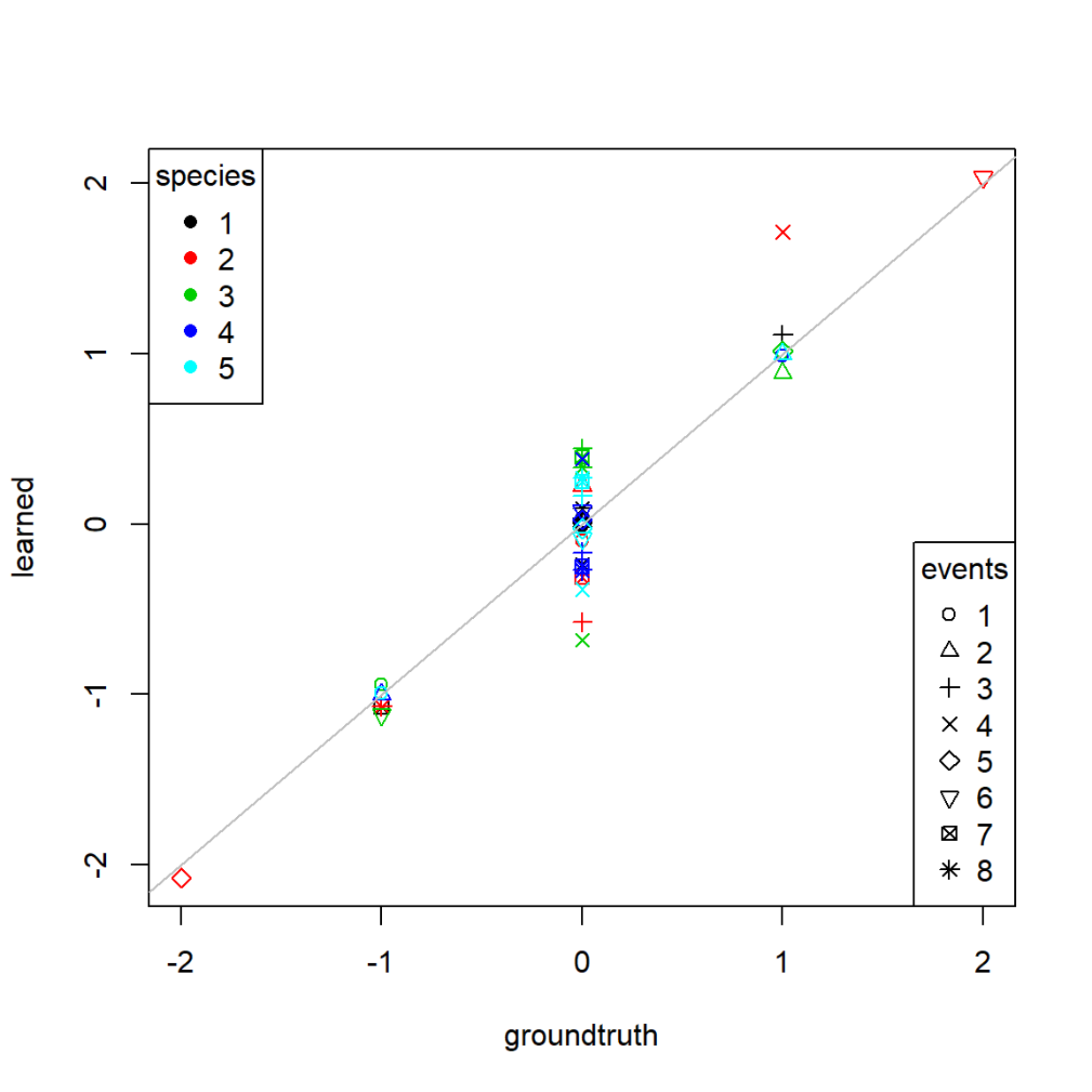}\hfill{}\includegraphics[width=0.24\columnwidth]{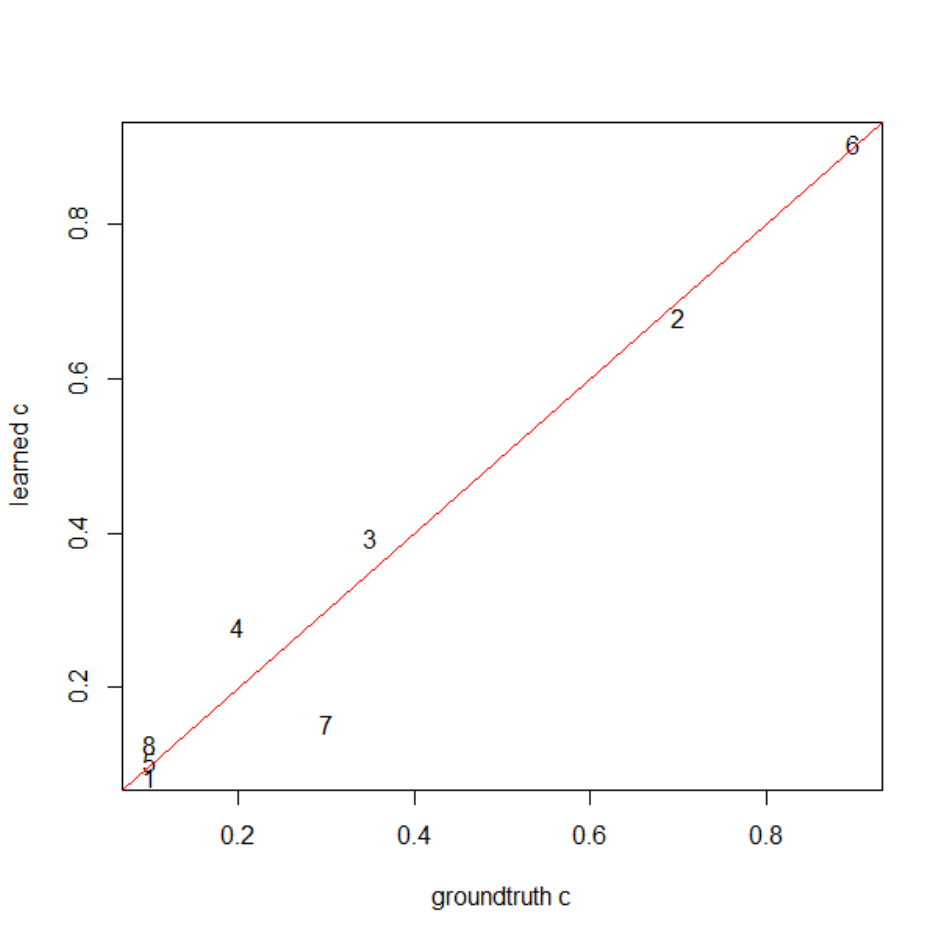}\hfill{}\includegraphics[width=0.24\columnwidth,height=0.24\columnwidth]{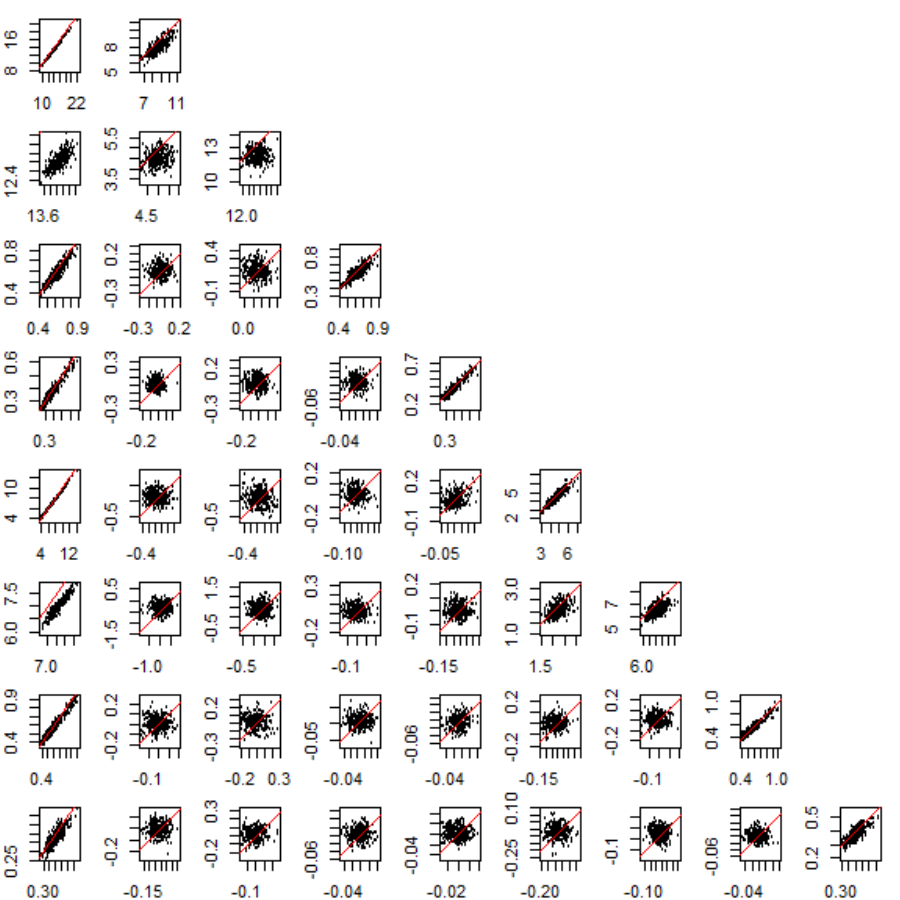}\hfill{}

\hfill{}(a) Stoichiometric matrix\hfill{}(b) S {*} h(x)\hfill{}(c) Rate constants\hfill{}(d) Event count mean/covariance\hfill{}

\caption{\label{fig:PAG}Comparing learned vs ground truth prokaryotic auto-regulation
dynamics.}
\end{figure}

Fig. \ref{fig:PAG} compares the typical learned prokaryotic auto-regulation
dynamics with the ground truth. We specify the event rates as $h_{v}(x)=c_{v}g_{v}(x)$,
where $v=1,\dots,8$, and $c_{v}$ is the event rate constant, and
compare the learned rate constants $c_{v}$ and stoichiometric matrix
$S$ element-wise with the ground truth (Fig. \ref{fig:PAG}a and Fig
\ref{fig:PAG}b). We also compare how the learned $S$ translates
event counts into state change (Fig. \ref{fig:PAG}c) and how the
learned event rate constants predict event statistics in one discrete
time step (Fig. \ref{fig:PAG}c) using a trajectory sampled from the
ground truth dynamics. The dynamics in a discrete time step is not
linear: In Fig. \ref{fig:PAG}(d), the mean and variance of event-1
count (subplots at row 1, column 1 and column 2) and event-5 count
(subplots at row 5, column 1 and column 6) are not equal, and the
event counts are correlated (columns 2-8, below the diagonal). Hence
we simulate the dynamics in sub-steps in the learning algorithm. While
a training data set with several thousands discrete time steps is
sufficient for the algorithm to learn to predict state-change statistics
between two time steps, a training data set with several million total
time steps that is comprised of many trajectories each starting from
a different initial state is required to learn the dynamics in Fig.
\ref{fig:PAG}.

\section{Conclusions}

In this paper, we formulated the problem of learning complex system dynamics as one of jointly learning a collection of local events whose rates (counts per unit time) are sparsely dependent the state variables and a stoichiometric matrix that mixes the event counts into the state variable changes. To learn a discrete-event representation of complex system dynamics, we showed that the back propagation algorithm to minimize the log evidence of a latent state dynamical system is a generalized expectation maximization algorithm. This approach is scalable because the graph representation of the model is fixed, and versatile because many real-world complex systems have discrete-event semantics. Our results show that the algorithm can data-efficiently capture complex network dynamics in several fields with meaningful events.

\section*{Broader Impact}
The research will achieve increased public scientific literacy and engagement through turning data into insights and civic engagement by enabling people to inspect how many complex systems run in detail. It will benefit our society through enabling people to see our society running at high-resolution and ask the what-if questions through developing a collection of machine learning tools and applications.

\bibliographystyle{plain}
\bibliography{cle}

\begin{thebibliography}{10}

\bibitem{alvarez2012kernels}
Mauricio~A Alvarez, Lorenzo Rosasco, Neil~D Lawrence, et~al.
\newblock Kernels for vector-valued functions: A review.
\newblock {\em Foundations and Trends{\textregistered} in Machine Learning},
  4(3):195--266, 2012.

\bibitem{battaglia2016interaction}
Peter Battaglia, Razvan Pascanu, Matthew Lai, Danilo~Jimenez Rezende, et~al.
\newblock Interaction networks for learning about objects, relations and
  physics.
\newblock In {\em Advances in neural information processing systems}, pages
  4502--4510, 2016.

\bibitem{battaglia2018relational}
Peter~W Battaglia, Jessica~B Hamrick, Victor Bapst, Alvaro Sanchez-Gonzalez,
  Vinicius Zambaldi, Mateusz Malinowski, Andrea Tacchetti, David Raposo, Adam
  Santoro, Ryan Faulkner, et~al.
\newblock Relational inductive biases, deep learning, and graph networks.
\newblock {\em arXiv preprint arXiv:1806.01261}, 2018.

\bibitem{bongard2007automated}
Josh Bongard and Hod Lipson.
\newblock Automated reverse engineering of nonlinear dynamical systems.
\newblock {\em Proceedings of the National Academy of Sciences},
  104(24):9943--9948, 2007.

\bibitem{bonilla2008multi}
Edwin~V Bonilla, Kian~M Chai, and Christopher Williams.
\newblock Multi-task gaussian process prediction.
\newblock In {\em Advances in neural information processing systems}, pages
  153--160, 2008.

\bibitem{borshchev2013big}
Andrei Borshchev.
\newblock {\em The big book of simulation modeling: multimethod modeling with
  AnyLogic 6}.
\newblock AnyLogic North America Chicago, 2013.

\bibitem{brunton2016discovering}
Steven~L Brunton, Joshua~L Proctor, and J~Nathan Kutz.
\newblock Discovering governing equations from data by sparse identification of
  nonlinear dynamical systems.
\newblock {\em Proceedings of the national academy of sciences},
  113(15):3932--3937, 2016.

\bibitem{chang2016compositional}
Michael~B Chang, Tomer Ullman, Antonio Torralba, and Joshua~B Tenenbaum.
\newblock A compositional object-based approach to learning physical dynamics.
\newblock {\em arXiv preprint arXiv:1612.00341}, 2016.

\bibitem{fox2015bayesian}
Emily~B Fox and David~B Dunson.
\newblock Bayesian nonparametric covariance regression.
\newblock {\em The Journal of Machine Learning Research}, 16(1):2501--2542,
  2015.

\bibitem{gillespie2007stochastic}
Daniel~T Gillespie.
\newblock Stochastic simulation of chemical kinetics.
\newblock {\em Annu. Rev. Phys. Chem.}, 58:35--55, 2007.

\bibitem{Golightly2007Diffusion}
A.~Golightly and D.~J. Wilkinson.
\newblock Bayesian inference for nonlinear multivariate diffusion models
  observed with error.
\newblock {\em Computational Statistics and Data Analysis}, 2007.

\bibitem{golightly2011bayesian}
Andrew Golightly and Darren~J Wilkinson.
\newblock Bayesian parameter inference for stochastic biochemical network
  models using particle markov chain monte carlo.
\newblock {\em Interface focus}, 1(6):807--820, 2011.

\bibitem{ma2022fewshot}
Chunwei Ma, Ziyun Huang, Mingchen Gao, and Jinhui Xu.
\newblock Few-shot learning via dirichlet tessellation ensemble.
\newblock In {\em International Conference on Learning Representations}, 2022.

\bibitem{marsan1994modelling}
Marco~Ajmone Marsan, Gianfranco Balbo, Gianni Conte, Susanna Donatelli, and
  Giuliana Franceschinis.
\newblock {\em Modelling with generalized stochastic Petri nets}.
\newblock John Wiley \& Sons, Inc., 1994.

\bibitem{matsim}
{MATSim development team (ed.)}.
\newblock {MATSIM-T: Aims, approach and implementation}.
\newblock Technical report, {IVT, ETH Z\"urich, Z\"urich}, 2007.

\bibitem{mattos2015recurrent}
C{\'e}sar Lincoln~C Mattos, Zhenwen Dai, Andreas Damianou, Jeremy Forth,
  Guilherme~A Barreto, and Neil~D Lawrence.
\newblock Recurrent gaussian processes.
\newblock {\em arXiv preprint arXiv:1511.06644}, 2015.

\bibitem{nodelman2002continuous}
Uri Nodelman, Christian~R Shelton, and Daphne Koller.
\newblock Continuous time bayesian networks.
\newblock In {\em Proceedings of the Eighteenth conference on Uncertainty in
  artificial intelligence}, pages 378--387. Morgan Kaufmann Publishers Inc.,
  2002.

\bibitem{opper2008variational}
Manfred Opper and Guido Sanguinetti.
\newblock Variational inference for markov jump processes.
\newblock In {\em Advances in Neural Information Processing Systems}, pages
  1105--1112, 2008.

\bibitem{o2006bayesian}
Anthony O’Hagan.
\newblock Bayesian analysis of computer code outputs: A tutorial.
\newblock {\em Reliability Engineering \& System Safety}, 91(10-11):1290--1300,
  2006.

\bibitem{pascanu2013construct}
Razvan Pascanu, Caglar Gulcehre, Kyunghyun Cho, and Yoshua Bengio.
\newblock How to construct deep recurrent neural networks.
\newblock {\em arXiv preprint arXiv:1312.6026}, 2013.

\bibitem{raissi2018deep}
Maziar Raissi.
\newblock Deep hidden physics models: Deep learning of nonlinear partial
  differential equations.
\newblock {\em The Journal of Machine Learning Research}, 19(1):932--955, 2018.

\bibitem{rao2013fast}
Vinayak Rao and Yee~Whye Teh.
\newblock Fast mcmc sampling for markov jump processes and extensions.
\newblock {\em Journal of Machine Learning Research}, 14(1):3295--3320, 2013.

\bibitem{ren2017clustering}
You Ren, Emily~B Fox, Andrew Bruce, et~al.
\newblock Clustering correlated, sparse data streams to estimate a localized
  housing price index.
\newblock {\em The Annals of Applied Statistics}, 11(2):808--839, 2017.

\bibitem{rudy2017data}
Samuel~H Rudy, Steven~L Brunton, Joshua~L Proctor, and J~Nathan Kutz.
\newblock Data-driven discovery of partial differential equations.
\newblock {\em Science Advances}, 3(4):e1602614, 2017.

\bibitem{schmidt2009distilling}
Michael Schmidt and Hod Lipson.
\newblock Distilling free-form natural laws from experimental data.
\newblock {\em science}, 324(5923):81--85, 2009.

\bibitem{teh2005semiparametric}
YW~Teh, M~Seeger, and MI~Jordan.
\newblock Semiparametric latent factor models.
\newblock In {\em AISTATS 2005-Proceedings of the 10th International Workshop
  on Artificial Intelligence and Statistics}, pages 333--340, 2005.

\bibitem{tran2017deep}
Dustin Tran, Rajesh Ranganath, and David~M Blei.
\newblock Deep and hierarchical implicit models.
\newblock {\em CoRR, abs/1702.08896}, 2017.

\bibitem{veestraeten2004conditional}
Dirk Veestraeten.
\newblock The conditional probability density function for a reflected brownian
  motion.
\newblock {\em Computational Economics}, 24(2):185--207, 2004.

\bibitem{wang2019towards}
Rui Wang, Karthik Kashinath, Mustafa Mustafa, Adrian Albert, and Rose Yu.
\newblock Towards physics-informed deep learning for turbulent flow prediction.
\newblock {\em arXiv preprint arXiv:1911.08655}, 2019.

\bibitem{wilkinson2011stochastic}
Darren~J Wilkinson.
\newblock {\em Stochastic modelling for systems biology}.
\newblock CRC press, 2011.

\bibitem{xu2016using}
Zhen Xu, Wen Dong, and Sargur~N Srihari.
\newblock Using social dynamics to make individual predictions: variational
  inference with a stochastic kinetic model.
\newblock In {\em Advances in Neural Information Processing Systems}, pages
  2783--2791, 2016.

\bibitem{zhang2017collapsed}
Boqian Zhang, Jiangwei Pan, and Vinayak~A Rao.
\newblock Collapsed variational bayes for markov jump processes.
\newblock In {\em Advances in Neural Information Processing Systems}, pages
  3749--3757, 2017.

\end{thebibliography}

\newpage
\section*{Supplementary Material}

\subsection*{Proof of Corollary \ref{cor:gradient} through Lemma \ref{thm:gradient-recursion}}

In this section, we explicitly show the relationship between the gradient-ascent approach and the expectation maximization approach to maximize the probability of observations of state-space process over a set of parameters. The goal is to maximize the log likelihood of the generative model $\log\sum_x p(x,y;\theta)$ with graduent $\mathbf E_{p(x|y)}\nabla_\theta\log p(x,y;\theta)$ where $x=\mathbf x_1,...,\mathbf x_N$ are latent variables, $y=\mathbf y_1,...,\mathbf y_N$ are observations, and $\theta$ are the parameters. E-M algorithm is to maximize $\mathbf E_{p(x|y;\theta_0)}\log p(x,y;\theta)$ over $\theta$ along the gradient $\mathbf E_{p(x|y;\theta_0)}\nabla_\theta\log p(x,y;\theta)$ until a maximum is identified, and then repeat the process with $\theta_0$ set to $\theta$ until a fixed point $\theta_0=\theta$ is reached. The gradient-ascent method doesn't require identifying an extremum point in maximizing $\mathbf E_{p(x|y;\theta_0)}\log p(x,y;\theta)$ over $\theta$ along the gradient $\mathbf E_{p(x|y;\theta_0)}\nabla_\theta\log p(x,y;\theta)$. In this sense, gradient ascent is a generalized E-M. 

\begin{replem}{thm:gradient-recursion}
For $n\le m$,
\begin{align*}
& \mathbf E_{p(\mathbf x_n|\mathbf y_{1},...,\mathbf y_m)}\nabla_\theta\log p(\mathbf x_n|\mathbf y_{1},...,\mathbf y_n)\ =\  \mathbf E_{p(\mathbf x_{n-1}|\mathbf y_{1},...,\mathbf y_m)}\nabla_\theta\log p(\mathbf x_{n-1}|\mathbf y_{1},...,\mathbf y_{n-1})\\
& \hspace{.4in}+  \mathbf E_{p(\mathbf x_{n-1},\mathbf x_n|\mathbf y_{1},...,\mathbf y_m)}\nabla_\theta\log p(\mathbf x_n, \mathbf y_n|\mathbf x_{n-1};\theta) -\nabla_\theta \log p(\mathbf y_n|\mathbf y_{1},...,\mathbf y_{n-1})
\end{align*}
\end{replem}

\begin{proof}[Proof of Lemma \ref{thm:gradient-recursion}]

The proof is based on moving the sum $\sum_{\mathbf x_{n-1}}$ out of the factor $\nabla\log p(\mathbf x_n|\mathbf y_1,...,\mathbf y_n)$, which is through $\nabla\log \sum_{\mathbf x_{n-1}} p=\frac{1}{\sum_{\mathbf x_{n-1}} p}\nabla \sum_{\mathbf x_{n-1}} p=\frac{1}{\sum_{\mathbf x_{n-1}} p}\sum_{\mathbf x_{n-1}}\nabla p=\sum_{\mathbf x_{n-1}}\frac{p}{\sum_{\mathbf x_{n-1}} p}\nabla\log p$, a trick to prove Theorem \ref{thm:gradient}. 
\begin{align*}
&\mathbf E_{p(\mathbf x_n|\mathbf y_1,...,\mathbf y_m)}\nabla\log p(\mathbf x_n|\mathbf y_1,...,\mathbf y_n) \\
= & \sum_{\mathbf x_n}p(\mathbf x_n|\mathbf y_1,...,\mathbf y_m)\nabla \log \frac{\sum_{\mathbf x_{n-1}} p(\mathbf x_{n-1}|\mathbf y_1,...,\mathbf y_{n-1}) p(\mathbf x_n,\mathbf y_n|\mathbf x_{n-1})}{p(\mathbf y_n|\mathbf y_1,...,\mathbf y_{n-1})}\\
= & \sum_{\mathbf x_n}\!\!\frac{p(\mathbf x_n|\mathbf y_1,...,\mathbf y_m)}{p(\mathbf x_n,\mathbf y_n|\mathbf y_1,...,\mathbf y_{n-1})}\nabla\!\!\sum_{\mathbf x_{n-1}}\!\! p(\mathbf x_{n-1}|\mathbf y_1,...,\mathbf y_{n-1}) p(\mathbf x_n,\mathbf y_n|\mathbf x_{n-1})\!\!-\!\! \nabla\!\log p(\mathbf y_n|\mathbf y_1,...,\mathbf y_{n-1})\\
= & \sum_{\mathbf x_n}\!\!\frac{p(\mathbf x_n|\mathbf y_1,...,\mathbf y_m)}{p(\mathbf x_n,\mathbf y_n|\mathbf y_1,...,\mathbf y_{n-1})}\!\!\sum_{\mathbf x_{n-1}}\!\!\nabla p(\mathbf x_{n-1}|\mathbf y_1,...,\mathbf y_{n-1}) p(\mathbf x_n,\mathbf y_n|\mathbf x_{n-1})\!\!-\!\! \nabla\!\log p(\mathbf y_n|\mathbf y_1,...,\mathbf y_{n-1})\\
= & \!\!\!\!\scriptstyle\sum\limits_{\mathbf x_{n-1},\mathbf x_n}\!\!\!\!\!\!\!\frac{p(\mathbf x_n|\mathbf y_1,...,\mathbf y_m) p(\mathbf x_{n-1}\mathbf x_n,\mathbf y_n|\mathbf y_1,...,\mathbf y_{n-1})}{p(\mathbf x_n,\mathbf y_n|\mathbf y_1,...,\mathbf y_{n-1})}\nabla\log p(\mathbf x_{n-1}|\mathbf y_1,...,\mathbf y_{n-1}) p(\mathbf x_n,\mathbf y_n|\mathbf x_{n-1}) -  \nabla \log p(\mathbf y_n|\mathbf y_1,...,\mathbf y_{n-1})\\
= & \!\!\!\!\scriptstyle\sum\limits_{\mathbf x_{n-1},\mathbf x_n} p(\mathbf x_n|\mathbf y_1,...,\mathbf y_m) p(\mathbf x_{n-1}|\mathbf x_n,\mathbf y_1,...,\mathbf y_{n})\nabla\log p(\mathbf x_{n-1}|\mathbf y_1,...,\mathbf y_{n-1}) p(\mathbf x_n,\mathbf y_n|\mathbf x_{n-1}) -  \nabla \log p(\mathbf y_n|\mathbf y_1,...,\mathbf y_{n-1})\\
= & \!\!\!\!\!\sum_{\mathbf x_{n-1},\mathbf x_n}\!\!\!\!\! p(\mathbf x_{n-1},\!\mathbf x_n|\mathbf y_1,...,\mathbf y_m)\!\nabla\!\log p(\mathbf x_{n-1}|\mathbf y_1,...,\mathbf y_{n-1}) p(\mathbf x_n,\mathbf y_n|\mathbf x_{n-1})\!\!-\!\!\nabla \log p(\mathbf y_n|\mathbf y_1,...,\mathbf y_{n-1})\\
= & \  \mathbf E_{p(\mathbf x_{n-1}|\mathbf y_{1},...,\mathbf y_m)}\nabla\log p(\mathbf x_{n-1}|\mathbf y_{1},...,\mathbf y_{n-1}) + \mathbf E_{p(\mathbf x_{n-1},\mathbf x_n|\mathbf y_{1},...,\mathbf y_m)}\nabla\log p(\mathbf x_n, \mathbf y_n|\mathbf x_{n-1}) \\
& \hspace{3.82in}- \nabla \log p(\mathbf y_n|\mathbf y_{1},...,\mathbf y_{n-1})
\end{align*}
In the above $p(\mathbf x_{n-1}|\mathbf x_n,\mathbf y_1,...,\mathbf y_{n}) = p(\mathbf x_{n-1}|\mathbf x_n,\mathbf y_1,...,\mathbf y_{m})$ due to Markov property.
\end{proof}

A corollary of the lemma is the following, because $\mathbf E_{p(\mathbf x_n|\mathbf y_{1},...,\mathbf y_n)}\nabla_\theta\log p(\mathbf x_n|\mathbf y_{1},...,\mathbf y_n)=0$ due to the property of scoring function:
\[\scriptstyle\nabla \log p(\mathbf y_n|\mathbf y_{1},...,\mathbf y_{n-1}) = \mathbf E_{p(\mathbf x_{n-1}|\mathbf y_{1},...,\mathbf y_n)}\!\!\nabla\log p(\mathbf x_{n-1}|\mathbf y_{1},...,\mathbf y_{n-1}) + \mathbf E_{p(\mathbf x_{n-1},\mathbf x_n|\mathbf y_{1},...,\mathbf y_n)}\!\!\nabla\log p(\mathbf x_n, \mathbf y_n|\mathbf x_{n-1};\theta).\]

\begin{repcor}{cor:gradient}Let $p(\mathbf x_{1},\dots,\mathbf x_{N},\mathbf y_{1},\dots,\mathbf y_{N};\theta)=\prod_{n}p(\mathbf x_{n}|\mathbf x_{n-1};\theta)p(\mathbf y_{n}|\mathbf x_{n};\theta)$ be the joint probability distribution of a latent process $\mathbf x_{1},...,\mathbf x_{N}$ and its observations $\mathbf y_{1},...,\mathbf y_{N}$ parameterized by $\theta$.
Then $\nabla_\theta\log p(\mathbf y_1,...,\mathbf y_N) = \sum_{n=1}^N\mathbf E_{p(\mathbf x_{n-1},\mathbf x_{n}|\mathbf y_1,...,\mathbf y_N)} \nabla_\theta\log p(\mathbf x_n,\mathbf y_n|\mathbf x_{n-1})$.
\end{repcor}

\begin{proof}[Proof of Corollary \ref{cor:gradient}]
\begin{align*}
 0 = & \mathbf{E}_{p(\mathbf x_{N}|\mathbf y_{1},...,\mathbf y_{N})}\nabla_{\theta}\log p(\mathbf x_{N}|\mathbf y_{1},...,\mathbf y_{N};\theta)\\
= & \sum\nolimits_{n=1}^{N}\mathbf{E}_{p(\mathbf x_{n-1},\mathbf x_{n}|\mathbf y_{1},...,\mathbf y_{N})}\nabla_{\theta}\log p(\mathbf x_{n},\mathbf y_{n}|\mathbf x_{n-1};\theta)-\nabla_{\theta}\log p(\mathbf y_{n}|\mathbf y_{1},...,\mathbf y_{n-1})\\
= & \sum\nolimits_{n=1}^N\mathbf E_{p(\mathbf x_{n-1},\mathbf x_n|\mathbf y_{1},...,\mathbf y_{N})} \nabla_\theta\log p(\mathbf x_n,\mathbf y_n|\mathbf x_{n-1}) - \nabla_{\theta}\log p(\mathbf y_{1},...,\mathbf y_{N})
\end{align*}
The first step, $\mathbf E_{p(\mathbf x_N|\mathbf y_{1},...,\mathbf y_{N})}\nabla_\theta \log p(\mathbf x_N|\mathbf y_{1},...,\mathbf y_{N};\theta)=0$, is due to the property of score function. The second step is the result of repeated application of Lemma \ref{thm:gradient-recursion}. The third step uses the Markov property of the joint distribution.
\end{proof}

\subsection*{Proof of Theorem~\ref{thm:existence}}

In this section, we give detailed proof of Theorem \ref{thm:existence} through building an Ito process $d \mathbf x_{t}=\mathbf \mu(\mathbf x_t) dt + \sqrt{\Sigma(\mathbf x_t)} d \mathbf B_t$ from the Langevian dynamics defined in Eq. \ref{eq:DSKM-state-model}.

\begin{repthm}{thm:existence}
Let $d \mathbf x_{t}=\mathbf \mu(\mathbf x_t) dt + \sqrt{\Sigma(\mathbf x_t)} d \mathbf B_t$, where $\mathbf  B_t$ is a Brownian motion process, $\mu(\mathbf x_t)$ is a drift term and $\Sigma(\mathbf x_t)$ is a covariance matrix. Let $\max_{\mathbf x}\|\operatorname{diag}(\sqrt{|\mu(\mathbf x)|})\|_2\le C_1$ and $\min_{\mathbf x}\|\sqrt{\Sigma(\mathbf x)}\|_2\ge C_2$. Then $\mathbf x_t$ can be written in the form of Langevian dynamics Eq. \ref{eq:DSKM-state-model}.
\end{repthm}

\begin{proof}[Proof of Theorem~\ref{thm:existence}]
We write $\mathbf x_t = \mathbf x_t^+ - \mathbf x_t^- $, where $\mathbf x_t^+=\max(\mathbf x_t, 0)$ and $\mathbf x_t^-=-\min(\mathbf x_t, 0)$. There is one-to-one correspondence between $\mathbf x_t$ and $\mathbf x_t^+$, $\mathbf x_t^-$ pairs.

Next we define populations as $(\mathbf z_t^+, \mathbf z_t^-)$, where $\mathbf z_t^+ = C \mathbf x_t^+$, $\mathbf z_t^- = C \mathbf x_t^-$, and $C$ is a constant satisfying $C\ge C_1/C_2$. The populations thus induce a new stochastic process $\mathbf z_t = \mathbf z_t^+ - \mathbf z_t^- = C \mathbf x_t$ with dynamics $d \mathbf z_{t}=C  \mu(\frac{1}{C}(\mathbf z_t^+-\mathbf z_t^-)) dt + \sqrt{C^2\Sigma(\frac{1}{C}(\mathbf z_t^+-\mathbf z_t^-))} d \mathbf B_t$, satisfying $C^2\Sigma-\operatorname{diag}(C|\mu|)$ is positive definite, because for all $\mathbf w\ne 0$, $\mathbf w^\top (C^2\Sigma) \mathbf w - \mathbf w^\top (C\operatorname{diag}|\mu|) \mathbf w \ge (C_2 C^2-C_1  C)\|\mathbf w\|_2^2>0$. The original process can then be expressed as two non-negative processes treated as populations, $\mathbf x_t = \frac{1}{C}(\mathbf z_t^+-\mathbf z_t^-)$. 

Next we define events each involving one population to account for the drift $C \mu(\mathbf x_t)$, and event pairs each involving two populations to account for the 
remaining diffusion $C^2\Sigma-\operatorname{diag}(C|\mu|)$. Specifically, we define the stoichiometric matrix in Eq. \ref{eq:SS}, where each column specifies a unique way (channel) for the populations to be changed through an event. Corresponding to each column in the stoichiometric matrix, we define the the aggregate event rates as elements of vector valued function $\mathbf h(\mathbf z^+_t,\mathbf z^-_t)$ in Eq. \ref{eq:hh}.  
\begin{align}
S =&\left[\begin{array}{cccccccc}
S_{0} & S_{\mu} & S_{{M \choose 2}}^{+} & S_{{M \choose 2}}^{-} & S_{{M \choose 2}}^{\pm} & S_{{M \choose 2}}^{\mp} & S_{\text{diag}}^{+} & S_{\text{diag}}^{-}\end{array}\right]\label{eq:SS}\\
\mathbf h =&\left[\begin{array}{cccccccc}
\mathbf h_{0}^\top & \mathbf h_{\mu}^\top & \mathbf h_{\Sigma^{+}}^\top & \mathbf h_{\Sigma^{+}}^\top & \mathbf h_{\Sigma^{-}}^\top & \mathbf h_{\Sigma^{-}}^\top & \mathbf h_{\text{diag}} & \mathbf h_{\text{diag}}^\top\end{array}\right]^\top,\label{eq:hh}
\end{align}

We specify the events below, where we use $\mathbb Z^{(i)+}$ and $\mathbb Z^{(i)-}$ to represent individuals belonging to the positive and negative populations $\mathbf z_t^+$ and $\mathbf z_t^-$. We use $f^+=\max(0,f)$ and $f^-=-\min(0,f)$ to represent the positive and negative part of $f$, $f$ is $\mu$ and $\Sigma_{i<j}$.

\begin{itemize}
    \item $S_0=\left[\begin{array}{c}
-I_{M\times M}\\
-I_{M\times M}
\end{array}\right]$ and $\mathbf h_{0}=\infty\cdot\mathbf{1}_{M\times1}$ correspond to $M$ events of the form $\mathbb Z^{(i)+}+\mathbb Z^{(i)-}\to\emptyset$, which cancels out a positive individual and a negative individual in the populations $\mathbf z_t^+$ and $\mathbf z_t^-$. $S_0$ is an $2M$-by-$M$ matrix while $\mathbf h_0$ is a $M$-dimensional non-negative function. This event happens infinitely fast. 
    \item $S_{\mu}=I_{2M\times2M}$ and $\mathbf h_{\mu}=\left[\begin{array}{c}
C\mu^{+}\\
C\mu^{-}
\end{array}\right]$ correspond to $M$ events of the form $\cdots\to\cdots+\mathbb Z^{(i)+}$, and another $M$ events of the form $\cdots\to\cdots+\mathbb Z^{(i)-}$. The effect is to account for the drift term $C \mu$.
    \item $S_{{M \choose 2}}^{+}=\left[\begin{array}{c}
\left(C_{2}^{M}\right)_{M\times\frac{M(M-1)}{2}}\\
0_{M\times\frac{M(M-1)}{2}}
\end{array}\right]$, $S_{{M \choose 2}}^{-}=\left[\begin{array}{c}
0_{M\times\frac{M(M-1)}{2}}\\
\left(C_{2}^{M}\right)_{M\times\frac{M(M-1)}{2}}
\end{array}\right]$, and $\mathbf h_{\Sigma^{+}}=\left[C^{2}\Sigma_{i<j}^{+}\right]$ correspond to $\frac{M(M-1)}{2}$ pairs of events of the form $\cdots\to\cdots+ \mathbb Z^{(i)+}+\mathbb Z^{(j)+}$ and $\cdots\to\cdots+ \mathbb Z^{(i)-}+\mathbb Z^{(j)-}$. The mean change to $z_t^{(i)}$ and $z_t^{(j)}$ is 0 and the effect is to account for the positive covariance $C^2 \Sigma_{i,j}$ by simultaneously increasing two random variables $z_t^{(i)}$ and $z_t^{(j)}$, and simultaneously decreasing them. The matrix $C_{2}^{M}=\left[\scalebox{0.75}{$\begin{array}{cccc|ccc|c|c}
1 & 1 & \cdots & 1 &  &  &  & \\
1 &  &  &  & 1 & \cdots & 1 & \\
 & 1 &  &  & 1 &  &  & \\
 &  & \ddots &  &  & \ddots &  & \ddots & 1\\
 &  &  & 1 &  &  & 1 &  & 1
\end{array}$}\right]$ selects two species in each column in different ways.
    \item $S_{{M \choose 2}}^{\pm}=\left[\scalebox{0.5}{$\begin{array}{cccc|ccc|c|c}
1 & 1 & \cdots & 1 &  &  &  & \\
 &  &  &  & 1 & \cdots & 1 & \\
 &  &  &  &  &  &  & \\
 &  &  &  &  &  &  & \ddots & 1\\
 &  &  &  &  &  &  & \\
\hline  &  &  &  &  &  &  & \\
1 &  &  &  &  &  &  & \\
 & 1 &  &  & 1 &  &  & \\
 &  & \ddots &  &  & \ddots &  & \ddots\\
 &  &  & 1 &  &  & 1 &  & 1
\end{array}$}\right]$, $S_{{M \choose 2}}^{\mp}=\left[\scalebox{0.5}{$\begin{array}{cccc|ccc|c|c}
 &  &  &  &  &  &  & \\
1 &  &  &  &  &  &  & \\
 & 1 &  &  & 1 &  &  & \\
 &  & \ddots &  &  & \ddots &  & \ddots\\
 &  &  & 1 &  &  & 1 &  & 1\\
\hline 1 & 1 & \cdots & 1 &  &  &  & \\
 &  &  &  & 1 & \cdots & 1 & \\
 &  &  &  &  &  &  & \\
 &  &  &  &  &  &  & \ddots\\
 &  &  &  &  &  &  &  & 1
\end{array}$}\right]$, and $h_{\Sigma^{-}}=\left[C^{2}\Sigma_{i<j}^{-}\right]$ correspond to $\frac{M(M-1)}{2}$ pairs of events of the form $\cdots\to\cdots+ \mathbb Z^{(i)+}+\mathbb Z^{(j)-}$ and $\cdots\to\cdots+ \mathbb Z^{(i)-}+\mathbb Z^{(j)+}$. The mean change to $z_t^{(i)}$ and $z_t^{(j)}$ is 0 and the effect is to account for the negative covariance $C^2 \Sigma_{i,j}$ by alternatively increasing and decreasing two random variables $z_t^{(i)}$ and $z_t^{(j)}$. The matrices $S_{{M \choose 2}}^{\pm}$ and $S_{{M \choose 2}}^{\mp}$ select one from the positive species and a different one from the negative species in different ways.
    \item $S_{\text{diag}}^{+}=\left[\begin{array}{c}
I_{M\times M}\\
0_{M\times M}
\end{array}\right]$, $S_{\text{diag}}^{-}=\left[\begin{array}{c}
0_{M\times M}\\
I_{M\times M}
\end{array}\right]$, and $h_{\text{diag}}=\left[C^{2}\Sigma_{i,i}-C\mu_{i}\right]$ are $M$ pairs of events of the form  $\cdots\to\cdots+\mathbb Z^{(i)+}$ and  $\cdots\to\cdots+\mathbb Z^{(i)-}$. The mean change to $z_t^{(i)}$ is 0 and the effect is to account for variance $C^2\Sigma_{i,i}-C\mu_i$ on the diagonal.
\end{itemize}

Next, we represent elements of $\mu$ and $\Sigma$ as their Taylor series expansions around 0 and in the radius of convergence: $\mu_i(\mathbf z_t^+,\mathbf z_t^-)=\sum_{|\alpha|\ge 0} (\mathbf z_t^+,\mathbf z_t^-)^\alpha\frac{\partial^\alpha\mu_i}{\alpha!}(0)=\sum_{|\alpha|\ge 0} (\mathbf z_t^+,\mathbf z_t^-)^\alpha(\frac{\partial^\alpha\mu_i^+}{\alpha!}(0)-\frac{\partial^\alpha\mu_i^-}{\alpha!}(0))$ and $\Sigma_{i,j}(\mathbf z_t^+,\mathbf z_t^-)=\sum_{|\alpha|\ge 0} (\mathbf z_t^+,\mathbf z_t^-)^\alpha\frac{\partial^\alpha\Sigma_{i,j}}{\alpha!}(0)=\sum_{|\alpha|\ge 0} (\mathbf z_t^+,\mathbf z_t^-)^\alpha(\frac{\partial^\alpha\Sigma_{i,j}^+}{\alpha!}(0)-\frac{\partial^\alpha\Sigma_{i,j}^-}{\alpha!}(0))$, where $\alpha$ is a multi-index. We identify each term in the Taylor series expansion as an event that contributes to one element of $\mathbf h$ in Eq. \ref{eq:hh} and corresponds to one column in Eq. \ref{eq:SS}.

The multiplicity of the reactants are multi-indexes $\alpha$, and the multiplicity of the products are $S+\alpha$ (Eqs. \ref{eq:production} and \ref{eq:S}). They are all non-negative. $S\mathbf h\Delta t=\mu_t\Delta t$, $S\operatorname{diag}\mathbf h\Delta t\cdot S^\top=\Sigma\Delta t$. By central limit theorem, as $\Delta t\to 0$, $\mathbf x_t$ equals the Langevian dynamics in Eq. \ref{eq:DSKM-state-model} in distribution.

\end{proof}

\subsection*{Learned Prokaryotic Auto-Regulation, and SynthTown Dynamics}

In this section, we report the learned individual level interactions in prokaryotic auto-regulation network and SynthTown traffic network from population-level observations. Code is included in the supplementary material and performance comparison with other learning algorithms is presented in the main text.

\end{document}